\documentclass[preprint,12pt]{elsarticle}

\usepackage{amssymb}
\usepackage{amsmath}
\usepackage[utf8]{inputenc}
\usepackage[T1]{fontenc}

\usepackage{array}      
\usepackage{tabularx}   
\usepackage{adjustbox}  
\usepackage{makecell}   

\usepackage{mathtools} 
 
\allowdisplaybreaks  
\usepackage{breqn} 
\usepackage{flexisym}
\usepackage{cuted} 
\usepackage{booktabs}   
\usepackage{graphicx}    
\usepackage{subcaption}  
\usepackage{float}       
\usepackage{adjustbox}   
\usepackage{lscape}
\usepackage{pdflscape}
\usepackage{subcaption}
\usepackage{enumitem}

\usepackage{multirow}

\usepackage[final]{changes}

\begin{document}

\begin{frontmatter}



\title{Genetic algorithm vs. gradient descent for training \replaced{a neural network architecture dedicated to low data regimes in}{a novel neural network architecture with low trainable parameter count for} small \deleted{imaging} medical datasets}


\author[inst1]{Amine Boukhari\textsuperscript{*}}
\author[inst2,inst3]{Boglarka Ecsedi}
\author[inst2]{Laszlo Papp}
\author[inst1]{Mathieu Hatt}

\affiliation[inst1]{organization={Laboratory of medical information processing (LaTIM), INSERM, UMR 1101,university of western brittany}, 
            }
\affiliation[inst2]{organization={Center for Medical Physics and Biomedical Engineering, Medical University of Vienna}, 
            }
\affiliation[inst3]{organization={Georgia Institute of Technology}, 
            }

\renewcommand*{\thefootnote}{\fnsymbol{footnote}}
\footnotetext[1]{* Corresponding author. Email: ihebamine@gmail.com}

\begin{abstract}
\textbf{Aim/Introduction} Distance-encoding biomorphic-informational neural network (DEBI-NN) is a recently proposed architecture in which connection weights are defined by the distances between neurons positioned in a Euclidian space. This approach drastically reduces the number of trainable parameters compared to classical neural networks in which weights are \added{directly} trained. The training process for DEBI-NN is based on a genetic algorithm (GA), rather than gradient descent (GD) which remains the prevailing optimization \replaced{algorithm}{method} in deep learning. We aim to design and implement a GD learner for DEBI-NN and assess its performance compared to GA.
\textbf{Materials and Methods} We designed a spatial backpropagation scheme tailored to DEBI-NN and carried out a comparison between GD and GA for classification tasks, using a synthetic non-linear “two-moons” dataset\replaced{,}{and} two clinical medical imaging radiomic datasets \added{and a fetal cardiotocography dataset} \added{with a sample sizes ranging from n=85 to n=2126}. Each optimizer was tuned through targeted hyperparameter searches adapted to each dataset.
\textbf{Results} Across all experiments, GA consistently produced superior decision boundaries and \replaced{classification}{generalization} performance (Synthetic: 100\% vs 83\%; DLBCL: 83\% vs 78\%; HECKTOR: 80\% vs 67\%; Fetal: 81\% vs 66\%), whereas GD exhibited instability and failed to fully capture the non-linear patterns inherent to DEBI-NN’s spatial encoding. The entangled gradients resulting from neuron interdependencies limit the effectiveness of classical backpropagation.
\textbf{Conclusion} These findings highlight fundamental limitations of gradient-based methods in architectures with highly interdependent spatial parameters and confirm the suitability of evolutionary strategies for training DEBI-NN. 
\end{abstract}

\begin{highlights}
\item A gradient-descent training algorithm tailored to the DEBI-NN architecture was designed.

\item Genetic Algorithms outperform Gradient Descent for DEBI-NN optimization.

\item Gradient descent struggles with the non-convex spatial encoding landscape.

\end{highlights}


\end{frontmatter}


\section{Introduction}
Over the last few years, the field of artificial intelligence (AI) has seen a significant increase in the size and complexity of neural network (NN) architectures \cite{llmscaling}\cite{sevilla2021parameter}. This is due to the availability of both a) larger training datasets and b) higher computational power. However, this is associated with a fast and exponential increase of the number of parameters that need to be trained in these new architectures. In the field of medical imaging \added{and signal}, data is often scarce for numerous applications (e.g., rare diseases, sub-types of specific cancer), rarely exceeding a sample size of several hundreds of patients. In addition, even when of relatively limited size, the datasets that can be collected are often highly heterogeneous due to the variability of scanning devices, acquisition protocols and reconstruction settings. The use of modern AI techniques including artificial NN (ANN), convolutional NN (CNN) and (vision) transformers (ViT) in the field of medical imaging has been of particular interest to address challenges such as automated segmentation of medical images, improved diagnosis and clinical decision making, as well as predictive modeling \cite{celard2023survey}\cite{takahashi2024comparison}. However, this has proved to be quite challenging to rely on such advanced and complex models due to the lack of availability of large datasets to train them properly for specific applications, such as rare diseases or specific cancer types. The use of techniques such as transfer learning \cite{raghu2019transfusion}, few-shot learning \cite{nayem2023few} or recent foundational models \cite{he2024foundation} fine-tuned to the application, generally fail to achieve their potential in clinical applications, mostly due to overfitting and a lack of generalizability.
Recently, a new NN architecture that drastically reduces the number of trainable parameters (down to a few percent depending on data and architecture) called Distance-encoding biomorphic-informational neural networks (DEBI-NN) has been proposed \cite{papp2023debi}. In this architecture, neurons are positioned in an euclidean space and their coordinates are trained, the weights linking the neurons to each other being defined by the distance after training, rather than training the weights directly as it is done in standard NNs.  This property results in a parameter count linearly proportional to the number of neurons (i.e., in case of a 3D space, 3 times the number of somas and axon terminals), whereas the parameter count increases polynomially in usual NNs models \cite{papp2023debi}. DEBI-NN is thus promising to allow training models benefitting from the strengths of NNs when only limited size datasets are available. It was shown that DEBI-NN can reach similar performance as NN in several tabular datasets for binary classification tasks \cite{papp2023debi}. Beyond the original DEBI-NN paper that demonstrated the parameter-count advantage over classical NNs, the concept of spatial plasticity and its benefits was further investigated in a recently published paper \cite{ECSEDI2025100008}. DEBI-NN spatial plasticity demonstrated intrinsic self-regularizing properties, allowing it to perform consistently better than classical NNs with less (or even no) regularization strategies, whereas classical NNs require a number of them to be combined in order to improve their performance. Nonetheless in the datasets used for comparison, even with these strategies implemented, classical NNs were still outperformed by DEBI-NN with no such strategy being used and no hyperparameter tuning, in a range of different numbers of hidden layers and their configurations \cite{ECSEDI2025100008}. These two initial investigations strongly imply that DEBI-NNs have a great potential to facilitate the use of NNs in small and/or imbalanced datasets, which are often dealt with in the field of medical imaging. In both these studies however, the learning of the DEBI-NN architecture was performed by relying  on a genetic algorithm (GA) rather than on the traditional gradient descent (GD) which is the state of the art for NNs\cite{atad2025neural(FE)}. This choice was made during the design and implementation of the DEBI-NN prototype, based on the intuitive assumption that gradient-descent would be ill-suited to the DEBI-NN architecture (see appendix section C.1 of \cite{papp2023debi}). However, it was not justified by a quantitative comparison with a properly designed and implemented GD adapted to it. Therefore, a formal comparison is required. In this work, we address this gap through two main contributions:
\begin{itemize}
    \item \textbf{The design of a backpropagation scheme tailored to DEBI-NN:} We propose a specific approach to overcome the difficulties posed by the architecture's non-convex optimization space, which makes classical backpropagation unsuitable.
    \item \textbf{The validation of the training strategy:} By evaluating this gradient-based approach against GA, we aim to confirm the suitability of the original evolutionary strategy, thereby validating the optimization choice for future studies relying on the DEBI-NN architecture.
\end{itemize}

The remainder of this paper is organized as follows. Section 2 describes the datasets, evaluation metrics, and implementation details of the proposed optimization strategies. Section 3 presents the experimental results and a comparative analysis across all considered datasets. Section 4 discusses the results and the implications and limitations of the proposed approach. Finally, Section 5 concludes the paper.

\section{Materials and Methods}
This section outlines the experimental framework established to compare the performance of Genetic Algorithms (GA) and Gradient Descent (GD) on the DEBI-NN architecture. We first describe the four datasets selected for this study (Section 2.1). We then detail the evaluation metrics used to assess classification performance (Section 2.2) and the implementation specifics, including the hyperparameter optimization strategy applied to both learners (Section 2.3). Finally, Section 2.4 presents a core contribution of this work: a modified backpropagation scheme specifically derived to adapt gradient descent to the spatial encoding constraints of the DEBI-NN architecture.

\subsection{Datasets}

\replaced{Four}{Three} datasets were relied on (Table~\ref{tab:datasets_summary}):
First, we used a simple "toy example" synthetic dataset to check the ability of the training processes to deal with non-linearity.
Second, we relied on two representative medical imaging datasets used for radiomics-based binary classification. Input data for DEBI-NN consisted of tabular sets combining clinical variables and image biomarker standardization initiative (IBSI) - compliant radiomic features (including geometrical shape, first-order intensity metrics, and higher-order textural features) extracted from 3D tumor volumes. These radiomics sets have been extracted previously for use in other research works. \added{Finally, we included an openly available fetal cardiotocography dataset\cite{cardiotocography_193}, consisting of measurements of fetal heart rate (FHR) and uterine contraction (UC) features, classified through a consensus of 3 expert obstetricians. This dataset was considered to assess the performance of the training processes on a multi-class and larger-scale dataset.}

\begin{itemize}
  \item \textbf{Synthetic Dataset}: A simple, non-linear 2D dataset designed to assess the capacity of both optimization algorithms to learn non-linear decision boundaries, An illustration of this dataset can be found in Figure~\ref{fig:synthetic_data}. The dataset comprises 800 training samples (406 from class 1 and 394 from class 2) and 200 test samples (106 from class 1 and 94 from class 2).
  
\begin{figure}[H]

  \centering
  \includegraphics[width=0.54\linewidth]{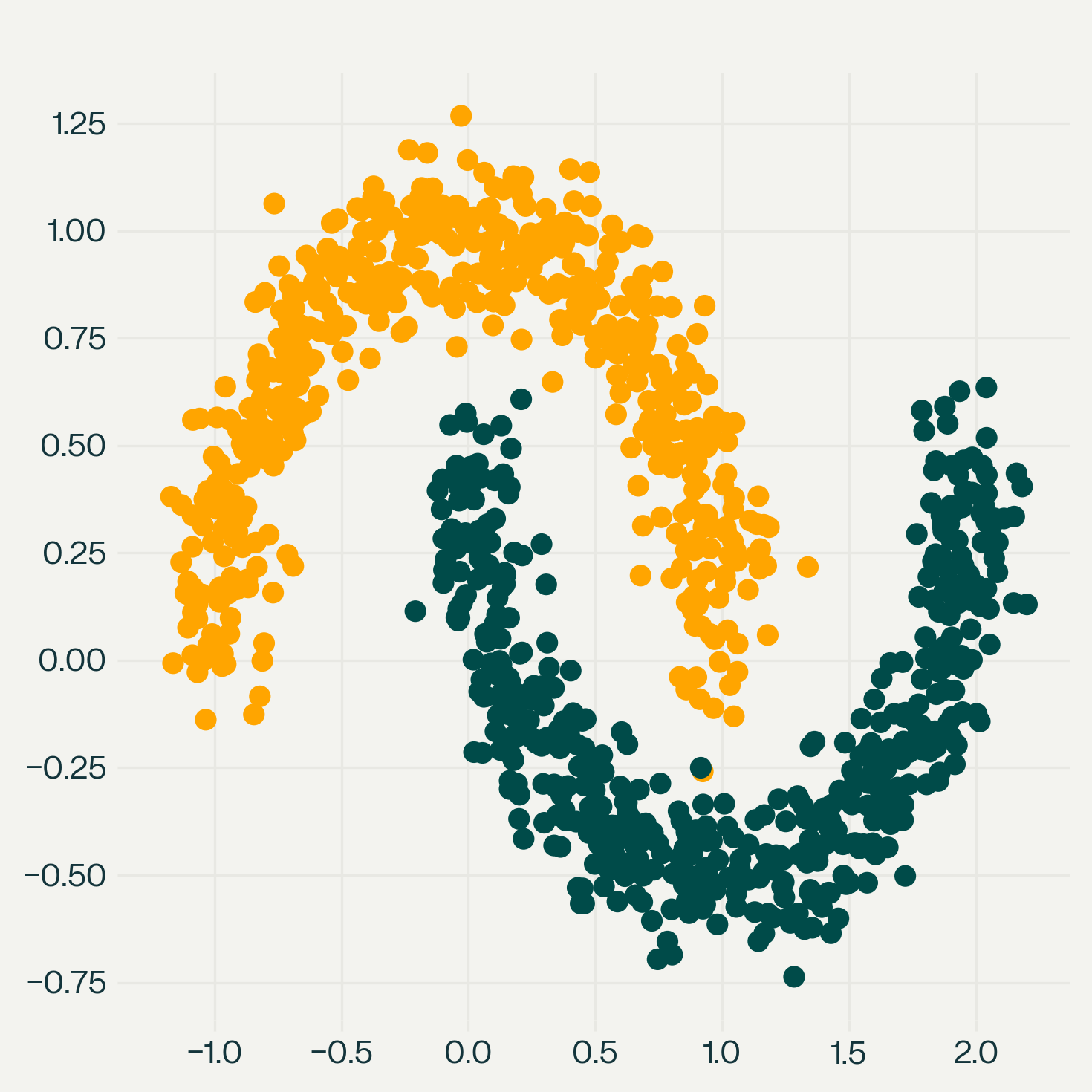}
  \caption{Synthetic 2D “two-moons” dataset used as toy example for non-linear classification.}
  \label{fig:synthetic_data}
\end{figure}

  \item \textbf{HECKTOR 2022 Dataset (HPV)}: This dataset was collected and curated within the context of the 2022 MICCAI challenge (HECKTOR) \cite{andrearczyk2022overview} \cite{Hecktor21} focused on segmentation and outcome prediction in head and neck cancer using 18F-FluoroDeoxyGlucose (FDG) Positron Emission Tomography/Computed Tomography (PET/CT) images. We used a subset of HECKTOR (the patients for which the information was available) for a binary classification task, namely the diagnosis of Human Papillomavirus (HPV) status. The input data combined 8 clinical variables (age, gender, stage, treatment, etc.) and 28 IBSI-compliant \cite{zwanenburg2020imageIBSI} 3D radiomic features (including geometrical shape metrics, first intensity based variables and higher order textural features) extracted from the delineated tumor volumes in both the FDG PET and CT scans. The dataset is notably heterogeneous, with images provided by several centers covering various PET/CT scanners types, acquisition protocols and reconstruction settings. It includes 158 training samples (99 HPV+ and 59 HPV-) and 74 test samples (55 HPV+ and 19 HPV-) . Examples of PET/CT images are provided in figure \ref{fig:hecktor_images}.

\begin{figure}[H]
  \centering
  \begin{subfigure}[b]{0.48\columnwidth}
    \centering
    \includegraphics[width=\textwidth]{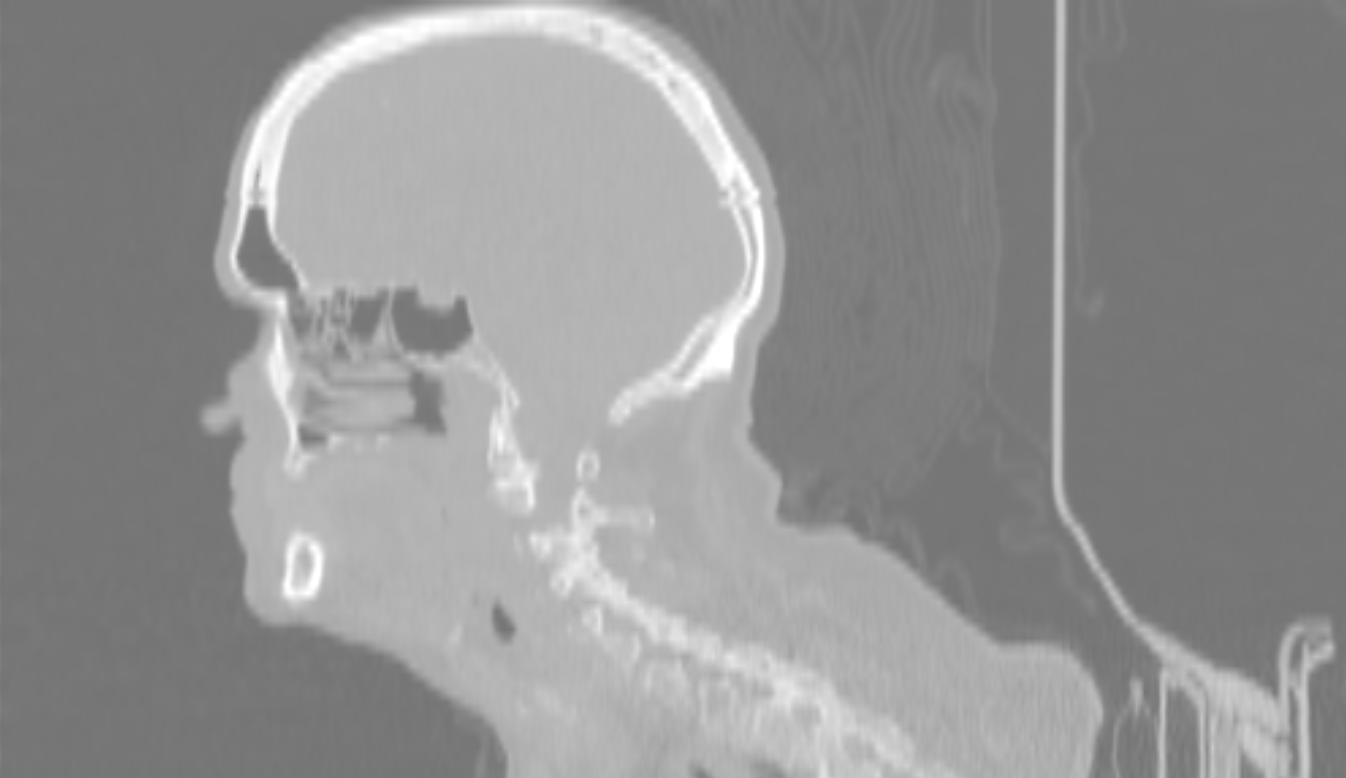}
    \caption{CT image}
    \label{fig:ct_hecktor}
  \end{subfigure}
  \hfill
  \begin{subfigure}[b]{0.48\columnwidth}
    \centering
    \includegraphics[width=\textwidth]{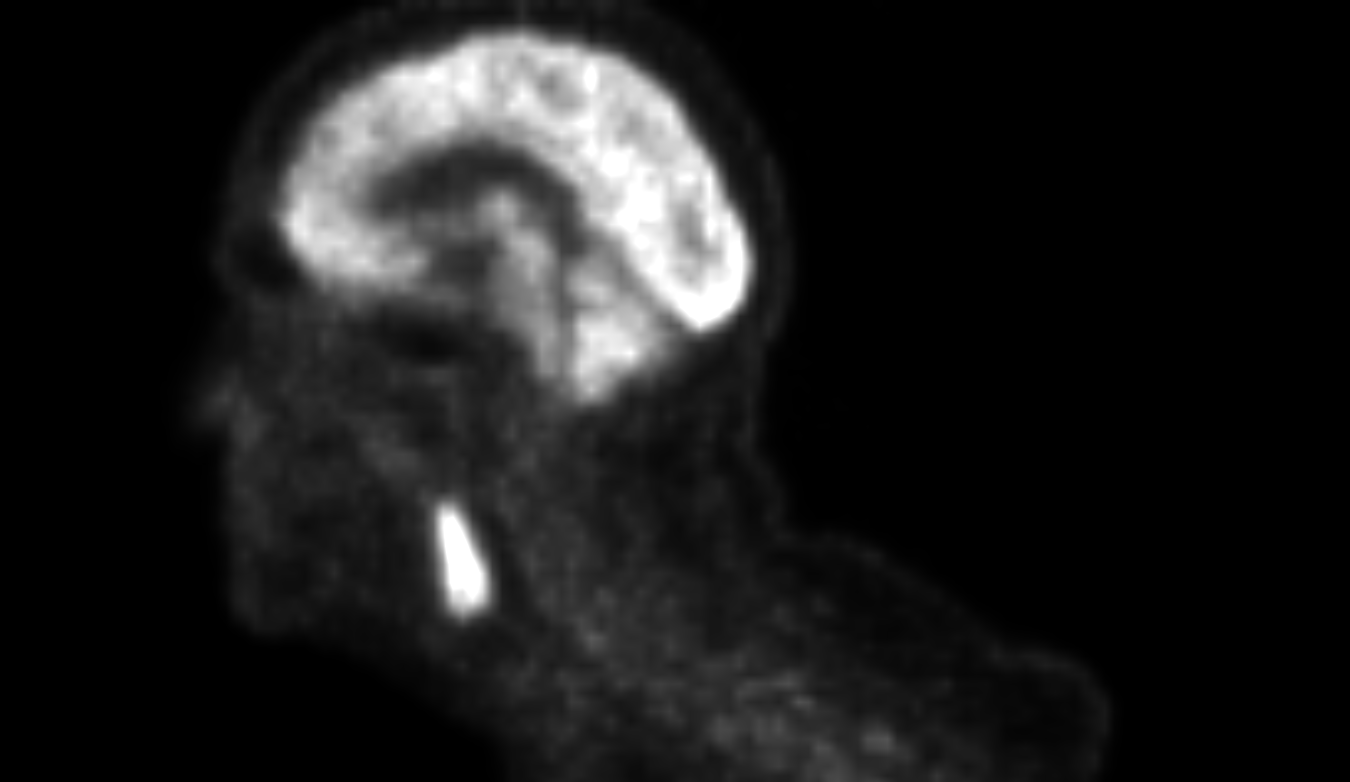}
    \caption{PET image}
    \label{fig:pet_hecktor}
  \end{subfigure}
  \caption{Example of PET and CT images from the HECKTOR dataset}
  \label{fig:hecktor_images}
\end{figure}

  \item \textbf{DLBCL Dataset (Diffuse Large B-Cell Lymphoma)\cite{ritter2022two}}: The binary classification task consists in predicting 2-year event-free survival (EFS). The tabular dataset contains 41 training cases (25 “no remission”, 16 “remission”) and 44 test cases (30 “no remission”, 14 “remission”), split based on the acquisition center.
  The dataset comprises both clinical data (7 features) and PET/CT scans from which 10 IBSI-compliant radiomic features were extracted. 

\item \textbf{\added{Fetal Cardiotocography Dataset} \cite{cardiotocography_193}}: \added{The task in this dataset consists in classifying the fetal state as Normal, Suspect, or Pathologic. The dataset contains 2126 fetal cardiotocograms (CTGs) characterized by 21 features, with labels obtained through a consensus of three expert obstetricians. The data is divided into a training set (1315 Normal, 245 Suspect, 140 Pathologic) and a test set (340 Normal, 50 Suspect, 36 Pathologic).}

\end{itemize}

\begin{table}[H]
  \centering
  \small
  \setlength{\tabcolsep}{4pt}      
  \renewcommand{\arraystretch}{1.2}

  \caption{Summary of the four datasets}
  \label{tab:datasets_summary}
  \begin{tabularx}{\columnwidth}{@{}%
      >{\raggedright\arraybackslash}p{0.12\columnwidth}  
      *{4}{>{\raggedright\arraybackslash}X}            
    @{}}
    \toprule
    Attribute                  & HECKTOR 2022                       & DLBCL            & \added{Fetal Cardiotocography}             & Synthetic (2D)                            \\
    \midrule
    Task                       & HPV status prediction (binary)             & 2-year event-free survival  (binary)& \added{Fetal state classification (3-class)} & Non-linear toy classification (binary)            \\
    \cmidrule(lr){2-5}
    Input Features             & Clinical + PET/CT radiomics        & Clinical + PET/CT radiomics      & \added{ Cardiotocography features }      & 2D coordinates                            \\
    \cmidrule(lr){2-5}
    \addlinespace[0.5em]  
    Samples (Train/Test)
                               & 158 / 74                           & 41 / 44       & \added{ 1700 / 426 }              & 800/200                                   \\
    \cmidrule(lr){2-5}
    Notes                      & Highly heterogeneous, multi-center & Split by acquisition center & \added{ Multi-class, larger-scale }& Noise-free; evaluates non-linear learning \\
    \bottomrule
  \end{tabularx}
\end{table}

\subsection{Evaluation\added{ metrics}}
In order to compare the performance of GA and GD learning processes for DEBI-NN, we report the \added{best classification performance in terms of} balanced accuracy (BAcc), sensitivity (Se) and specificity (Sp) reached for by both approaches \added{(table \ref{tab:ga_gd_bacc})}.
Note that each result is the best performance after searching for optimized hyperparameters in each learning \replaced{algorithm}{method} (which may differ).

\added{We also provide the mean, median, standard deviation, minimum and maximum values of BAcc obtained across the range of hyperparameters configurations tested, table 3. Note that the number of configurations tested for hyperparameters are not the same for GA and GD, as we had to explore alternative options previously disregarded for GA (see implementation section below). The distributions were statistically compared with the Mann-Whitney U test and p-values are reported in table 3. All performance metrics and details about all hyperparameters configurations explored are provided in the supplemental material.}

\subsection{Implementation \added{details}}

For each training session, multiple parameter sets were tested in order to identify
the optimal configuration for each dataset and each optimization strategy.
The parameters yielding the best test results are reported in the Results section in table \ref{tab:hyperparams_part1} and \ref{tab:hyperparams_part2}. The full set of hyperparameters used in the study is available at Mendeley data \cite{Boukhari2025DEBINNData}.

\added{
The choice of \added{hyper}parameters was \added{partly} guided by the results of the study on the impact of regularization methods on DEBI-NN \cite{ECSEDI2025100008}, particularly for GA.}

\added{
For GD, given the limited knowledge about the most suitable hyperparameters, due to differences from classical network implementations and the specific nature of the DEBI-NN involving novel parameters, a wider range of hyperparameters was explored.}

\added{The varied parameters include:}

\begin{itemize} 
  \item \textbf{Learning rate (lr)}
  \item \textbf{Number of neurons}, with the value range adjusted based on the dataset
  \item \textbf{Number of layers}, between 1 and 3
  \item \textbf{Weight Standardization}
  \item \textbf{Initial neuron placement}, chosen among:
    \begin{itemize}
      \item \emph{Random coordinates}: similar to Random weight initialization in classical neural networks
      \item \emph{Onion}: neurons are arranged in nested spherical layers to facilitate information accessibility
      \item \emph{Singularity}: all neurons are initialized at the same location; applicable only with the genetic algorithm
    \end{itemize}
  \item \textbf{Distance‐to‐weight mapping function}, selected from the following:
    \begin{itemize}
      \item \emph{Gaussian}
      \item \emph{Inverse}: weight = $1 - \dfrac{\mathrm{distance}}{\mathrm{maxdistance}}$
    \end{itemize}
\end{itemize}

\added{A more detailed description of the DEBI-NN hyperparameters can be found in the handbook \cite{papp_2025_17224628}. In total, 6 to 10 configurations were explored for GA and 16 to 25 for GD.}

\subsubsection{Genetic Algorithm Baseline}

The Genetic Algorithm (GA) baseline employed in this study utilizes the original DEBI-NN implementation described in \cite{papp2023debi}, where a more detailed description can be found. This approach relies on evolutionary principles modeling natural selection, crossover, and mutation to optimize the model parameters. It encodes the 3D spatial coordinates $(x, y, z)$ of somas and axons as genes within a digital chromosome, thereby fully defining the network's spatial structure.

The training process operates iteratively through the following steps:
\begin{itemize}
    \item \textbf{Initialization and Selection:} An initial population of network variants is randomly generated. At each iteration, parents are chosen via tournament selection, prioritizing individuals that minimize the cross-entropy loss.
    \item \textbf{Reproduction:} Offspring are generated by combining genes from two parents with equal probability (50\% crossover). Subsequently, these genes undergo random mutations, enabling the exploration of new spatial configurations.
\end{itemize}

The final model is selected as the best individual in the population. This strategy allows the network to spatially evolve toward an optimal distribution of distances, which are subsequently mapped to connection weights.

\subsubsection{\added{Loss function:}}
\added{
The loss function $\mathcal{L}$ is defined as the average Weighted Cross-Entropy loss over a batch of $N$ samples:
}

\[
\mathcal{L} = \frac{1}{N} \sum_{k=1}^{N} \mathcal{L}_k
\]
\added{where the loss for a single sample $\mathcal{L}_k$ is given by:}
\[
\mathcal{L}_k = - \sum_{c=1}^{M} \beta_c \cdot y_{k,c} \cdot \log(\hat{p}_{k,c})
\]
\added{Where:}
\begin{itemize}
    \item \added{$N$ is the total number of samples in the batch.}
    \item \added{$M$ is the total number of classes.}
    \item \added{$\mathcal{L}_k$ is the loss for the $k$-th sample.}
    \item \added{$\beta_c$ is the weight assigned to class $c$.}
    \item \added{$y_{k,c}$ is true if the given prediction is correct for the given sample.}
    \item \added{$\hat{p}_{k,c}$ is the model's predicted probability.}
\end{itemize}

\subsubsection{\added{Rationale for Gradient Computation}}

\added{
During the development of the gradient-based version, the question came up whether the gradient should be computed with respect to the weights or the distances between axons and somas.}

\added{Computing the gradient on the distances presents a drawback: DEBI-NN operates directly on distances and applies a mapping function to convert these distances into weights only when inference is required \cite{papp2023debi}. This mapping is non-linear, and the standard gradient descent algorithm was not originally designed to handle such a case.}

\added{On the other hand, computing the gradient on the weights has its own limitation. Since DEBI-NN fundamentally relies on distances, this approach would introduce additional computational complexity, as it would require repeatedly converting between distances and weights at each optimization step.}

\added{The adopted solution was to compute the gradient with respect to the distances while introducing a linear mapping function corresponding to the inverse function previously described in this paper. This resolves the limitations of the distance-based approach while preserving consistency with DEBI-NN’s design.}

\subsubsection{GroupNorm \added{Backpropagation} Implementation}
Gradient descent optimization is prone to instability issues, frequently manifesting as either vanishing \cite{abuqaddom2021oriented(vanishing_gradient)}\cite{liu2023improved(vanishing_autoencoder)} or exploding gradients. However, in the specific case of DEBI-NN, preliminary results revealed a pronounced tendency towards the latter, with frequent gradient explosion when using gradient descent. This motivated the implementation of Group Normalization\cite{wu2018groupnormalization}, chosen for its independence from batch size, making it compatible with potential future convolutive versions of DEBI-NN.
As the implementation details of GroupNorm \added{Backpropagation} in fully connected networks were lacking in the literature, we present below the mathematical formulation used in our implementation.

\vspace{\baselineskip}
\textbf{Output layer :}
\begin{subequations}%
\begin{align*}
\frac{\partial C}{\partial w}
 &= \frac{\partial C}{\partial \hat{y}}
   \frac{\partial \hat{y}}{\partial a}
   \frac{\partial a}{\partial z}
   \frac{\partial z}{\partial w}
  = e\,\gamma\,\sigma'(z)\,\hat{y}_{L-1} \\[6pt]
\frac{\partial C}{\partial\beta}
 &= \frac{\partial C}{\partial \hat{y}}
   \frac{\partial \hat{y}}{\partial\beta}
  = e \\[6pt]
\frac{\partial C}{\partial\gamma}
 &= \frac{\partial C}{\partial \hat{y}}
   \frac{\partial \hat{y}}{\partial\gamma}
  = e\,\hat{a}
\end{align*}
\end{subequations}

\textbf{Hidden layers :}
\begin{subequations}%
\begin{align*}
\frac{\partial C}{\partial\gamma} &= e\,\hat{a} \\[6pt]
\frac{\partial C}{\partial\beta} &= e \\[6pt]
\frac{\partial C}{\partial w}
 &= \frac{\partial C}{\partial \hat{y}}
   \frac{\partial \hat{y}}{\partial \hat{a}}
   \frac{\partial \hat{a}}{\partial a}
   \frac{\partial a}{\partial z}
   \frac{\partial z}{\partial w} \\[2pt]
 &= e\,\gamma\,
   \frac{\partial \hat{a}}{\partial a}\,
   \sigma'(z)\,\hat{y}_{L-1}
\end{align*}
\end{subequations}

\textbf{Derivative of $\hat{a}(i)$} with respect to $a(i)$
\begin{multline*}
\frac{\partial \hat{a}(i)}{\partial a(i)}
 = \frac{\frac{\partial (a(i)-\mu)}{\partial a(i)} \sqrt{\delta^{2}+\epsilon}
     - (a(i)-\mu) \frac{\partial \sqrt{\delta^{2}+\epsilon}}{\partial a(i)}}%
    {\delta^{2}+\epsilon}
\\[6pt]
 = \frac{\left(1-\frac{1}{m}\right)\sqrt{\delta^{2}+\epsilon}
     - (a(i)-\mu)
      \left[
       \frac{1}{2\sqrt{\delta^{2}+\epsilon}}
       \left(
        \frac{2}{m}\left(1-\frac{1}{m}\right)(a(i)-\mu)
        -\frac{2}{m}\sum_{\substack{j=1\\j\neq i}}^{m}
         \frac{1}{m}\left(a^{(j)}-\mu\right)
       \right)
      \right]}%
    {\delta^{2}+\epsilon}
\end{multline*}

\textbf{Legend:}
\begin{align*}
C & : \text{Cost function (loss)} \\
w & : \text{Network weight} \\
\hat{y} & : \text{Final output of the neuron (} \hat{y} = \hat{a} \cdot \gamma + \beta \text{)} \\
\hat{y}_{L-1} & : \text{Final output of a neuron from the previous layer (L-1)} \\
a & : \text{Activation value (output of } \sigma \text{, before normalization)} \\
\hat{a} & : \text{Normalized activation} \\
z & : \text{Weighted input to the neuron (pre-activation)} \\
\sigma'(z) & : \text{Derivative of the activation function applied to } z \\
\mu & : \text{Mean of the group of activations} \\
\delta^2 & : \text{Variance of the group of activations} \\
\epsilon & : \text{Small constant for numerical stability} \\
m & : \text{Number of elements in the group (GroupNorm group size)} \\
e & : \text{Error term, the incoming gradient } \left( \frac{\partial C}{\partial \hat{y}} \right) \\
\gamma, \beta & : \text{Learnable scaling and shifting parameters in GroupNorm} \\
L & : \text{Index of the current layer} \\
i & : \text{Index of the current neuron in the group}
\end{align*}

\textbf{Notation Convention}
\begin{itemize}
  \item \textbf{Lowercase letters} ($a, w, z, \beta, \gamma$): Used for scalar values.
  \item \textbf{Uppercase letters} ($C$): Used for standalone concepts like the Cost function.
\end{itemize}

\subsection{\replaced{Spatial Backpropagation for DEBI-NN}{Adapting Backpropagation to DEBI-NN}}

\begin{figure} [H]
  \centering
  \includegraphics[width=\columnwidth]{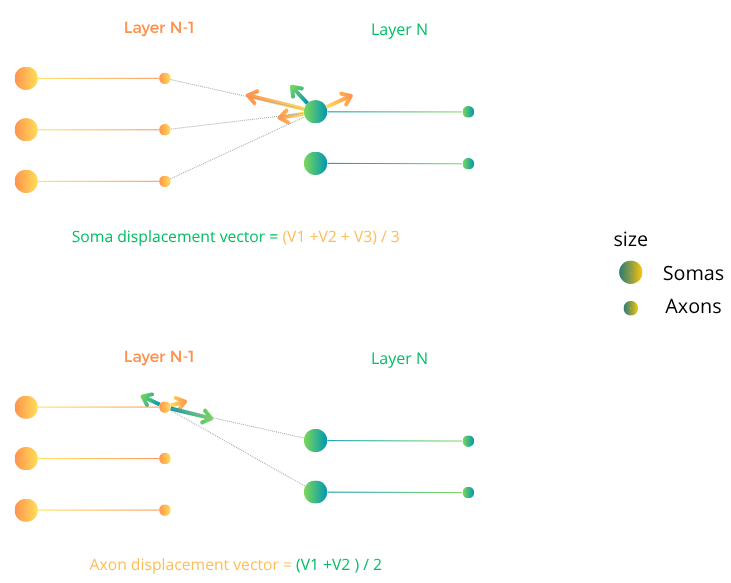}
  \caption{Illustration of soma (large circles) and axon (small circles) displacement vectors during gradient-based optimization in DEBI-NN.
Each Axon or Soma receives gradient-induced displacement vectors from its connected neurons. The soma displacement (top) is computed as the average of all incoming vectors. The axon displacement (bottom) is similarly updated based on connected somas from the next layer. This reflects the spatial propagation of gradients in the 3D geometry of DEBI-NN.
}
  \label{fig:displacement vector}
\end{figure}

Unlike conventional NNs in which weights are learned directly, the DEBI-NN architecture defines weights through the spatial distances between neurons positions in the Euclidian space. As for the original DEBI-NN publication, we implemented all models in the 3D Euclidian space, however the proposed scheme could also be considered with a different number of dimensions.
Consequently, the parameters to be optimized in the training process are the neurons’ 3D coordinates. This raises a specific challenge, as modifying a single neuron's position affects all connection distances to and from that neuron, which breaks the assumption of weight independence that is made in gradient backpropagation.
To resolve this, we redefine the gradient update rule as follows: for each connection, a displacement vector is computed \added{from the backpropagated gradients} by differentiating the loss with respect to the distance, and then projecting this gradient into a spatial direction from one neuron to the other. Each neuron thus accumulates multiple such displacement vectors, one per connection. The final update for a given neuron is the average of these vectors, effectively reconciling potentially conflicting gradients. This \replaced{algorithm}{method} ensures that neuron positions are adjusted in a globally consistent way, maintaining DEBI-NN’s structural integrity during training. Figure \ref{fig:displacement vector} illustrates the process.

An important point is that soma and axon coordinates are not updated in the same optimization step. Changing either endpoint already alters every distance and therefore every weight; applying the same gradient twice (first to somas and then to axons) would use a no longer valid information for the second update.
Thus, We alternate updates in two phases:

\begin{enumerate}[label=\textbf{(\roman*)}]
  \item At iteration \(t\) (even), all axon coordinates are kept frozen, forward and backward passes are computed, and the displacement vectors are applied only to the somas.
  \item At iteration \(t{+}1\) (odd), the forward and backward passes are recomputed with the updated geometry and the new displacement vectors are applied only to the axons.
\end{enumerate}

\section{Results}
The results obtained across all hyperparameter configurations tested for this study are available at mendeley data \cite{Boukhari2025DEBINNData}. Table \ref{tab:ga_gd_bacc} reports the best performance metrics for GA and GD across all four datasets and table \ref{tab:ga_gd_stats} reports the balanced accuracy statistics obtained across the entire range of tested hyperparameter configurations.

Tables \ref{tab:ga_gd_bacc} and \ref{tab:ga_gd_stats} show that GA consistently outperformed GD across the \added{four} datasets, both in terms of the best-performing model and in the distributions of BAcc across all tested hyperparameter configurations. 
It is important to acknowledge that a wider range of hyperparameters was explored for GD, inherently increasing the variance and the likelihood of observing lower performance compared to GA, for which suboptimal configurations were filtered out in prior studies. However, restricting the statistical comparison to comparable configurations yielded similar statistical trends.

\begin{figure}[htbp]
    \centering
    
    \begin{subfigure}[b]{0.49\linewidth}
        \centering
        \includegraphics[width=\linewidth]{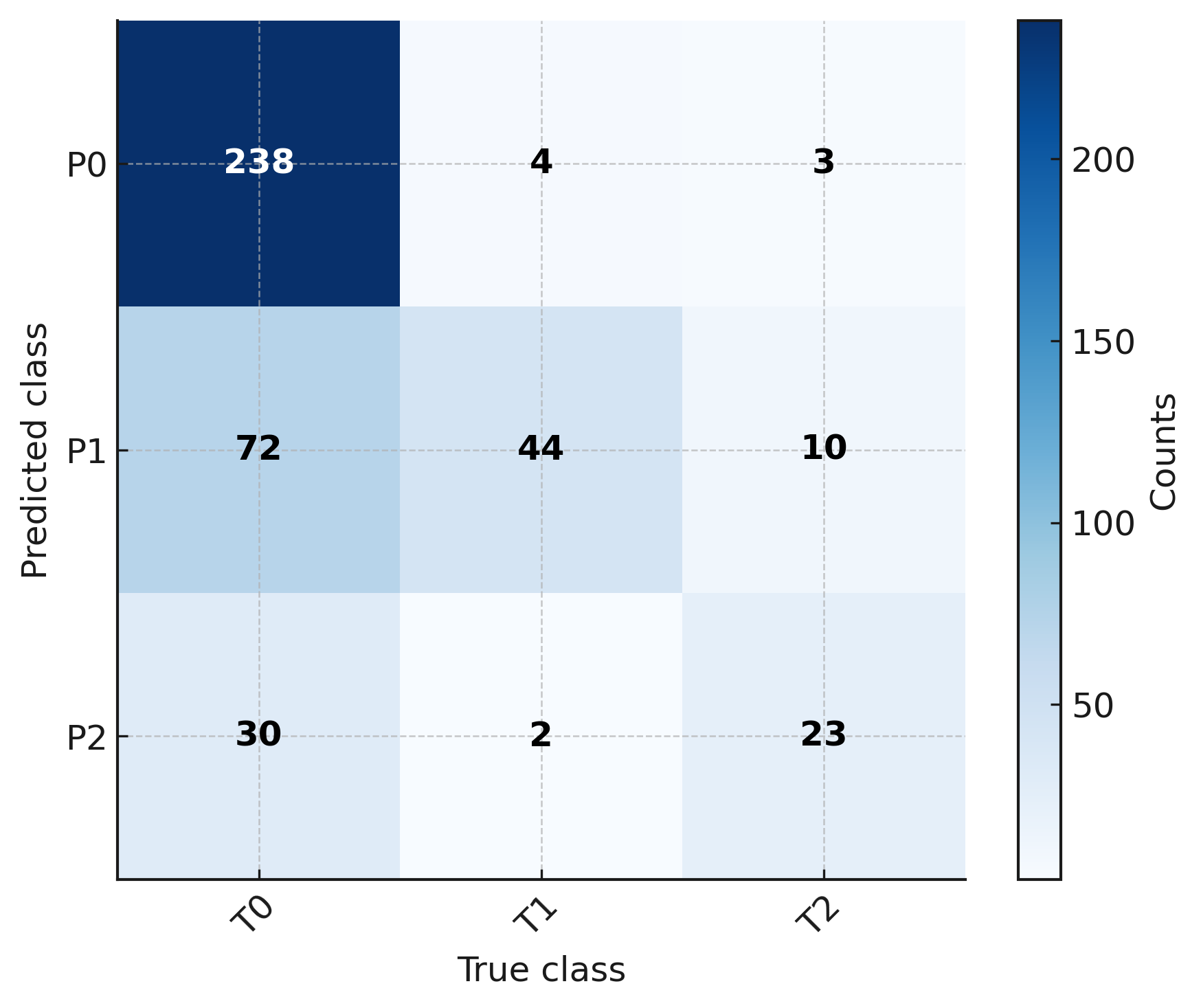}
        \caption{\added{\textbf{GA} Confusion matrix}} 
        \label{fig:confusion_matrix1}
    \end{subfigure}
    \hfill
    \begin{subfigure}[b]{0.49\linewidth}
        \centering
        \includegraphics[width=\linewidth]{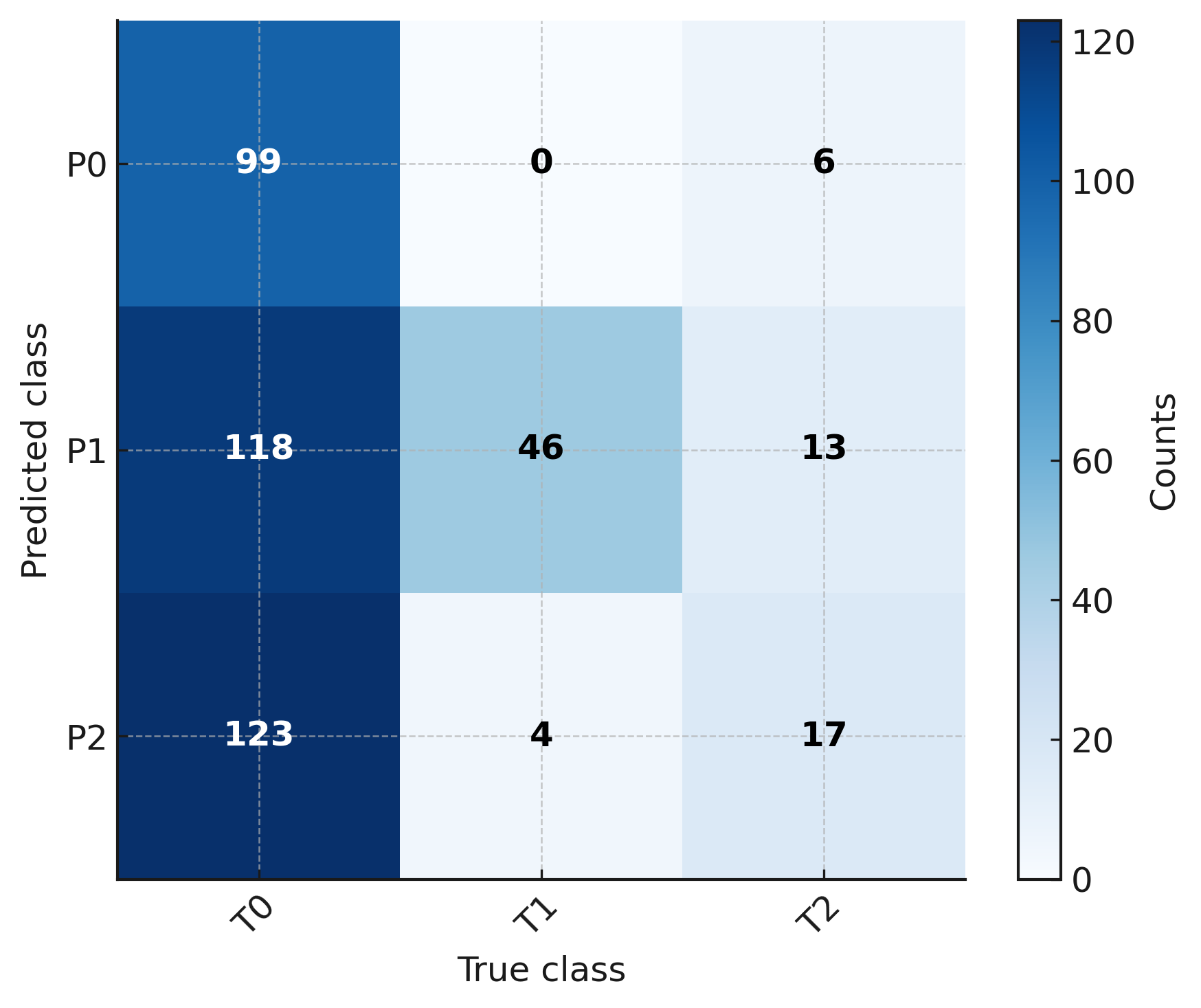}
        \caption{\added{\textbf{GD} Confusion matrix}}
        \label{fig:confusion_matrix2}
    \end{subfigure}
    
    \vspace{4mm} 

    \caption{\added{Confusion \textbf{matrices} for the best performing models on the fetal cardiotocography dataset.}} 
    \label{fig:confusion-matrix-fetal}
\end{figure}

\begin{figure}[H]
  \centering

  \begin{subfigure}[b]{0.6\columnwidth}
    \centering
    \includegraphics[width=\linewidth]{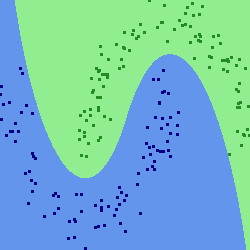}
    \caption{Genetic optimization on the synthetic dataset \added{(BAcc: 100\%). The misclassified region represents 17.46\% of the total area.}}

    \label{fig:genetic_sub}
  \end{subfigure}

  \vskip 0.5em

  \begin{subfigure}[b]{0.6\columnwidth}
    \centering
    \includegraphics[width=\linewidth]{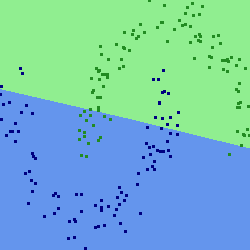}
    \caption{Gradient-based optimization on the synthetic dataset \added{(BAcc: 83\%). The misclassified region represents 24.32\% of the total area.}}
    \label{fig:gradient_sub}
  \end{subfigure}

  \caption{Learned decision boundaries using genetic and gradient-based optimization on the synthetic dataset.}
  \label{fig:optimization_boundaries}
\end{figure}

In terms of the best obtained results with optimized hyperparameters configurations (i.e., the most relevant comparison between GA and GD, table \ref{tab:ga_gd_bacc}), on the synthetic two-moons task, GA achieved perfect BAcc (100\%), whereas GD plateaued at 83\% (84\% sensitivity, 82\% specificity), indicating GD’s inability to fully capture non-linear decision boundaries, as confirmed by the decision-boundary plot in Figures \ref{fig:genetic_sub} and \ref{fig:gradient_sub} and misclassification rate representing respectively 18 and 24 percents of the surface. \added{A similar trend was observed} on the PET/CT radiomics binary classification tasks, where GA yielded 83\% \added{BAcc} versus 78\% for GD in the DLBCL cohort and 80\% \added{BAcc} versus 66\% in HECKTOR, with GA reaching perfect specificity (100\%) in both the synthetic and HECKTOR datasets. Finally, the same trend was observed for the multi-class Fetal Cardiotocography dataset, where GA achieved 81\% BAcc (74\% sensitivity, 87\% specificity) compared to 66\% for GD (56\% sensitivity, 75\% specificity). The confusion matrices (figure \ref{fig:confusion-matrix-fetal}) show that GA yielded significantly fewer misclassified cases. Specifically, GD primarily struggled to distinguish between classes 1 and 2 (118 errors vs. 72 for GA), and between classes 1 and 3 (123 errors vs. 30 for GA).

When comparing performance of GA and GD over the entire range of hyperparameters combinations (6 to 10 configurations for GA and 16 to 25 configurations for GD, table \ref{tab:ga_gd_stats}), GA obtained higher BAcc compared to GD for all 4 datasets, with statistical trends showing low p-values (from 0.0017 to 0.0631). The best BAcc were obtained in the synthetic datasets, with GD having lower mean (74\% vs. 88\%) and nearly double the standard deviation (14\% vs 8.8\%) compared to GA, p=0.0017. In contrast, the lowest performance was obtained on the 3-class classification task in the fetal dataset, where GA obtained slightly higher BAcc (64\% vs 57\%) with a slightly larger standard deviation (9\% vs 5\%), p=0.0348. Note however that the difference in terms of best performance was much higher (max value of 81\% for GA, versus 67\% only for GD). For the two radiomics datasets, the difference in performance in DLBCL was the lowest (70\% vs. 65\%, with close standard deviations of 7.4\% and 7.8\%, p=0.0631), a trend also observed for the best BAcc (83\% vs. 78\%). The difference was larger for HECKTOR (p=0.0025), where GA obtained a mean BAcc of 66\%, GD reaching only 54\%. However the standard deviation of GA was larger (11\% vs. 7\%). This also corresponds to a large difference in the best result (80\% for GA vs. only 67\% for GD).

\begin{table}[H]
  \centering
  \small
  \setlength{\tabcolsep}{4pt}
  \renewcommand{\arraystretch}{1.1}
  \caption{\added{Best} performance metrics for GA and GD \added{across all datasets}}
  \label{tab:ga_gd_bacc}
  \begin{tabular}{@{}l c c c c@{}}
    \toprule
    Metric                      & Synthetic & DLBCL & HECKTOR & \added{Fetal Cardiotocography} \\
    \midrule
    GA Balanced Accuracy (\%)   & 100       & 83    & 80   & \added{81}\\
    GD Balanced Accuracy (\%)   &  83       & 78    & \replaced{67 }{66}   & \added{66}\\
    \cmidrule(lr){2-5}
    GA Sensitivity (\%)         & 100       & 86    & 60   & \added{74}\\
    GD Sensitivity (\%)         &  84       & 79    & \replaced{49}{58}   & \added{56}\\
    \cmidrule(lr){2-5}
    GA Specificity (\%)         & 100       & 80    & 100  & \added{87}\\
    GD Specificity (\%)         &  82       & 77    & \replaced{84}{74}   & \added{75}\\
    \bottomrule
  \end{tabular}
\end{table}

\begin{table}[H]
  \centering
  \small
  \setlength{\tabcolsep}{4pt}
  \renewcommand{\arraystretch}{1.1}
  \caption{\added{BAcc performance across all runs for GA and GD}}
  \label{tab:ga_gd_stats}
  \begin{tabular}{@{}l c c c c@{}}
    \toprule
    \added{Metric} & \added{Synthetic} & \added{DLBCL} & \added{HECKTOR} & \added{Fetal Cardiotocography} \\
    \midrule
    \added{GA Mean ±  std. deviation  (\%)}& \added{88 ± 8.8 }& \added{70 ±  7.4}& \added{66 ± 11.1}& \added{64 ± 8.81}\\
    \added{GD Mean ±  std.deviation (\%)}& \added{74 ±  14}& \added{65 ±  7.8}& \added{54 ±  7.1}& \added{57 ±  5.41}\\
    \cmidrule(lr){2-5}
 \added{GA Median (\%)}& \added{83}& \added{67} & \added{70} & \added{61}\\
 \added{GD Median (\%)}& \added{80}& \added{65}& \added{53}& \added{59}\\
    \cmidrule(lr){2-5}
    \added{GA [min, max] (\%)}& \added{[81, 100]}& \added{[58, 83]}& \added{[50, 80]}& \added{[54, 81]}\\
    \added{GD [min, max](\%)}& \added{[26, 83]}& \added{[48, 78]}& \added{[42, 67]}& \added{[50, 67]}\\
    \cmidrule(lr){2-5}
    \added{p-value}  & \added{0.0017} & \added{0.0631} & \added{0.0025}& \added{0.0348} \\
    \bottomrule
  \end{tabular}
\end{table}

\begin{table}[H]
  \centering
  \footnotesize
  \caption{Optimal hyperparameter settings (Part 1): Architecture and Training}
  \label{tab:hyperparams_part1}
  \begin{adjustbox}{max width=\textwidth}
    \begin{tabular}{ll cccc}
      \toprule
      \textbf{Dataset} & \textbf{Optimizer} & \textbf{Neurons per layer} & \textbf{Initialization} & \textbf{Mapping func.} & \textbf{Epochs} \\
      \midrule
      \multirow{2}{*}{Synthetic} & GA & 16,16 & Singularity & Gaussian & 5000 \\
                                 & GD & 16,16 & Random & Inverse & 250 \\
      \cmidrule(l){2-6}
      \multirow{2}{*}{DLBCL}     & GA & 85,64,48 & Singularity & Gaussian & 1000 \\
                                 & GD & 16,16,16 & Random & Gaussian & 350 \\
      \cmidrule(l){2-6}
      \multirow{2}{*}{HECKTOR}   & GA & 180 & Singularity & Gaussian & 1000 \\
                                 & GD & 72,72 & Random & Gaussian & 1000 \\
      \cmidrule(l){2-6}
      \multirow{2}{*}{Fetal}     & GA & 128,128 & Singularity & Gaussian & 400 \\
                                 & GD & 128,128 & Onion & Inverse & 400 \\
      \bottomrule
    \end{tabular}
  \end{adjustbox}
\end{table}

\begin{table}[H]
  \centering
  \footnotesize
  \caption{Optimal hyperparameter settings (Part 2): Regularization and Optimizer-specifics}
  \label{tab:hyperparams_part2}
  \begin{adjustbox}{max width=\textwidth}
    \begin{tabular}{ll cccc}
      \toprule
      \textbf{Dataset} & \textbf{Optimizer} & \textbf{Groupnorm} & \textbf{Weight Stand.} & \textbf{L1/L2} & \textbf{Population count} \\
      \midrule
      \multirow{2}{*}{Synthetic} & GA & On & Off & 0 / 0 & 5000 \\
                                 & GD & On & Off & 0 / 0 & - \\
      \cmidrule(l){2-6}
      \multirow{2}{*}{DLBCL}     & GA & On & On & 0 / 0 & AUTO \\
                                 & GD & On & Off & 0.05 / 0.05 & - \\
      \cmidrule(l){2-6}
      \multirow{2}{*}{HECKTOR}   & GA & On & On & 0 / 0 & AUTO \\
                                 & GD & On & On & 0 / 0 & - \\
      \cmidrule(l){2-6}
      \multirow{2}{*}{Fetal}     & GA & On & Off & 0.01 / 0.01 & AUTO \\
                                 & GD & On & Off & 0.01 / 0.01 & - \\
      \bottomrule
    \end{tabular}
  \end{adjustbox}
\end{table}

\begin{figure}[htbp]
    \centering
    
    
    \begin{subfigure}[b]{0.49\linewidth} 
        \centering
        \includegraphics[width=\linewidth]{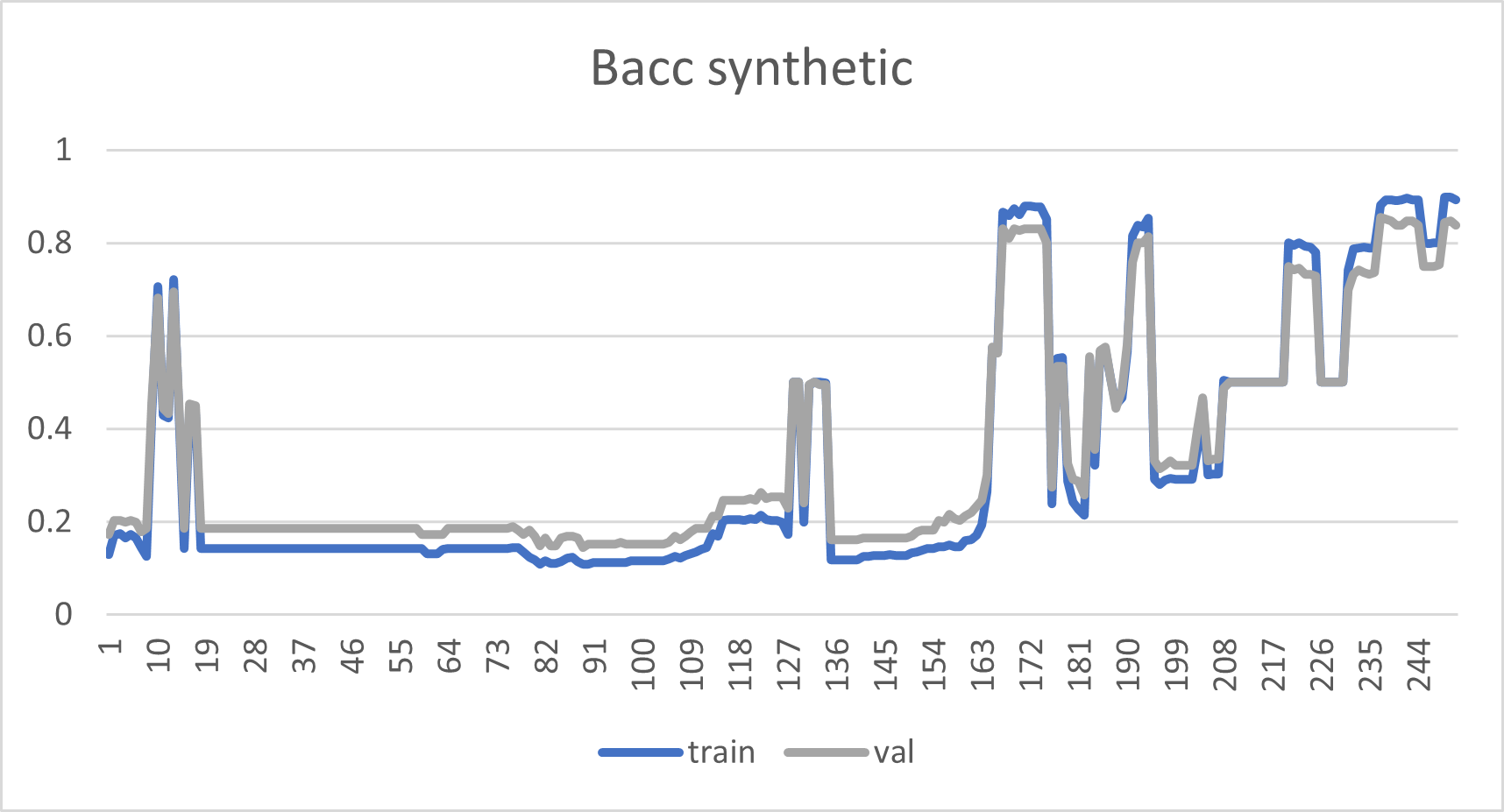}
        \caption{\added{Balanced Accuracy Curve on the \textbf{Synthetic} Dataset}}
        \label{fig:sub1}
    \end{subfigure}
    \hfill
    \begin{subfigure}[b]{0.49\linewidth} 
        \centering
        \includegraphics[width=\linewidth]{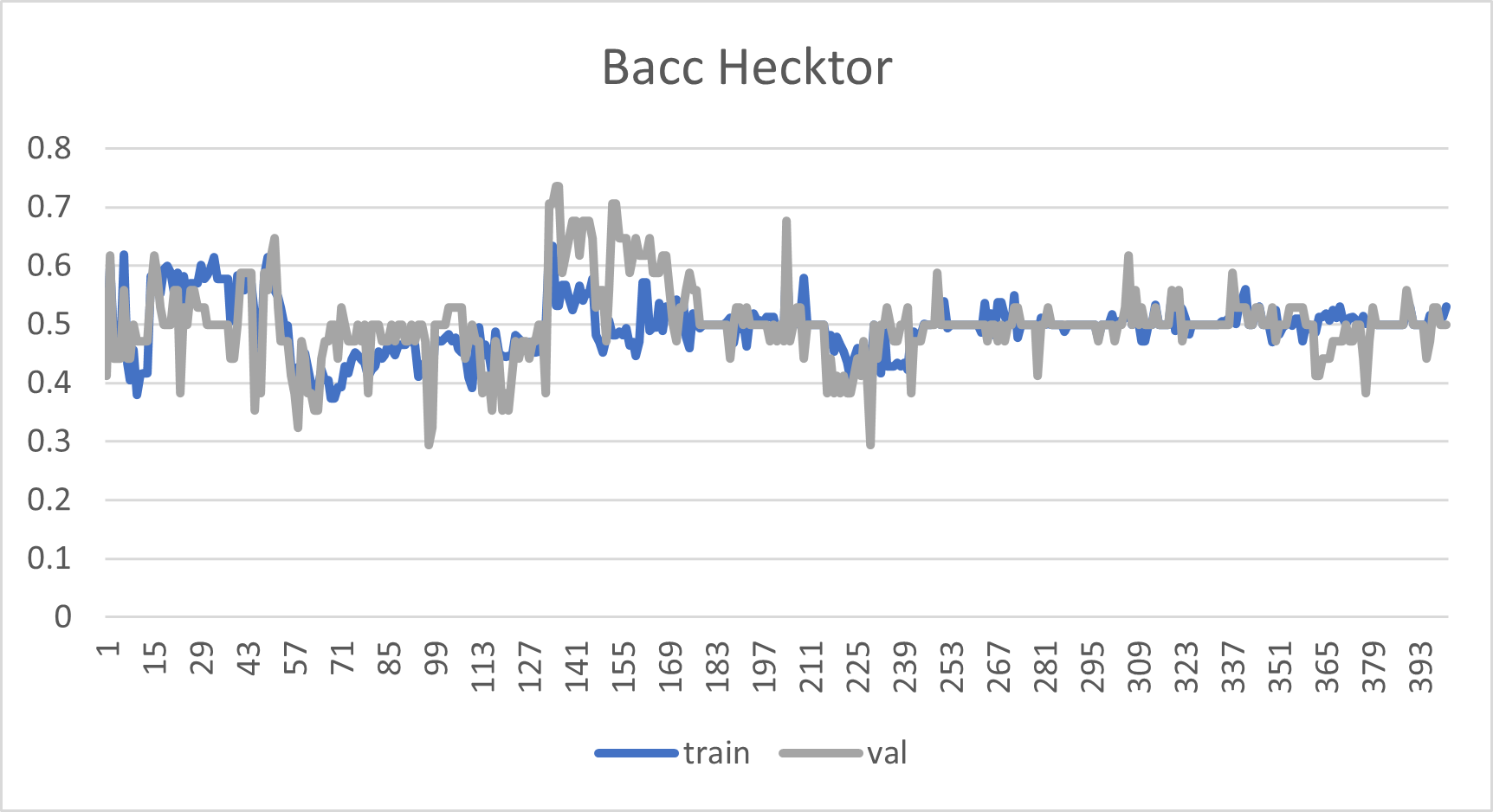}
        \caption{\added{Balanced Accuracy Curve on the \textbf{Hecktor} Dataset}}
        \label{fig:sub2}
    \end{subfigure}
    
    \vspace{1em}
    
    
    \begin{subfigure}[b]{0.49\linewidth} 
        \centering
        \includegraphics[width=\linewidth]{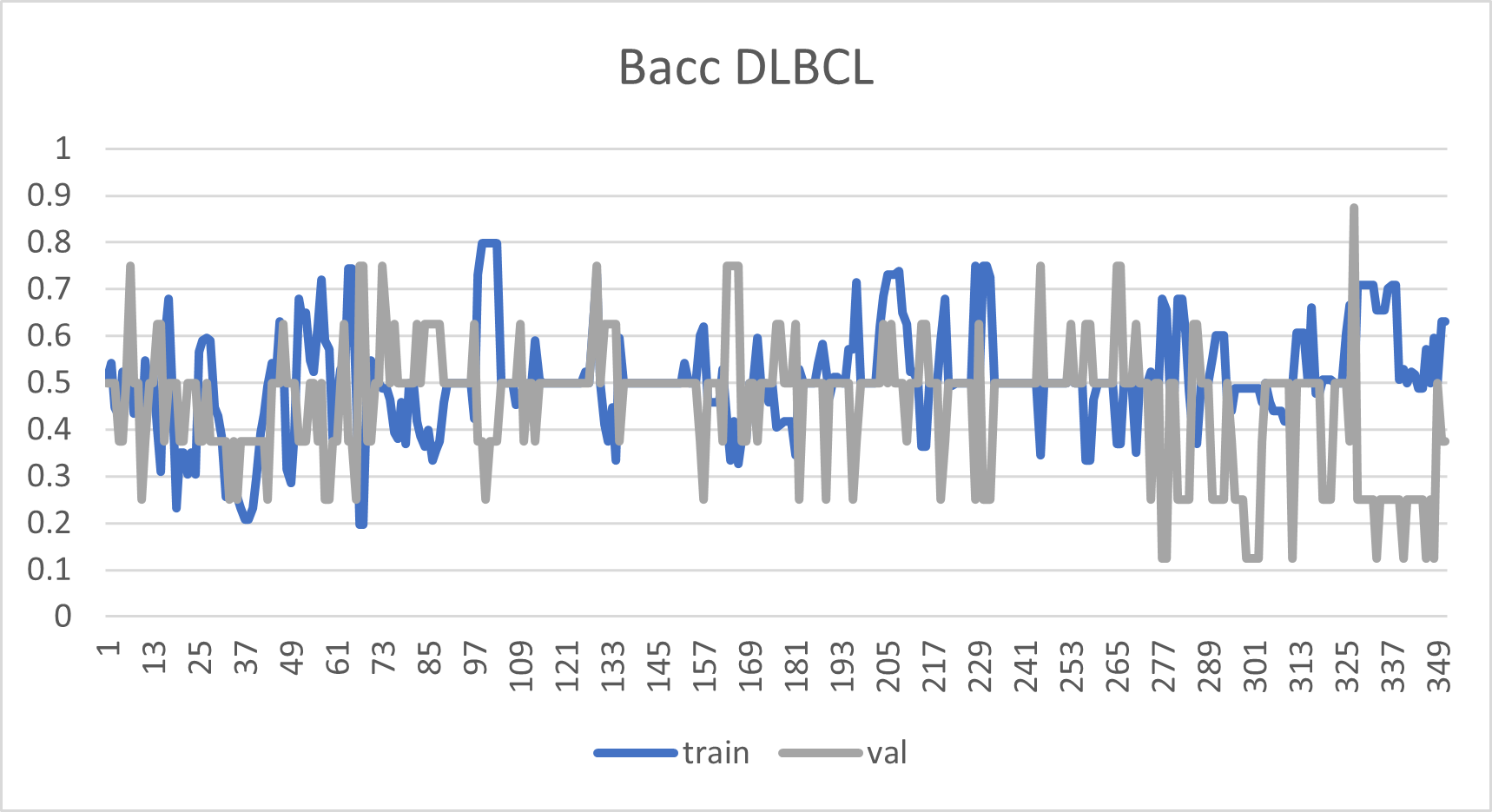}
        \caption{\added{Balanced Accuracy Curve on the \textbf{DLBCL} Dataset}}
        \label{fig:sub3}
    \end{subfigure}
    \hfill
    \begin{subfigure}[b]{0.49\linewidth} 
        \centering
        \includegraphics[width=\linewidth]{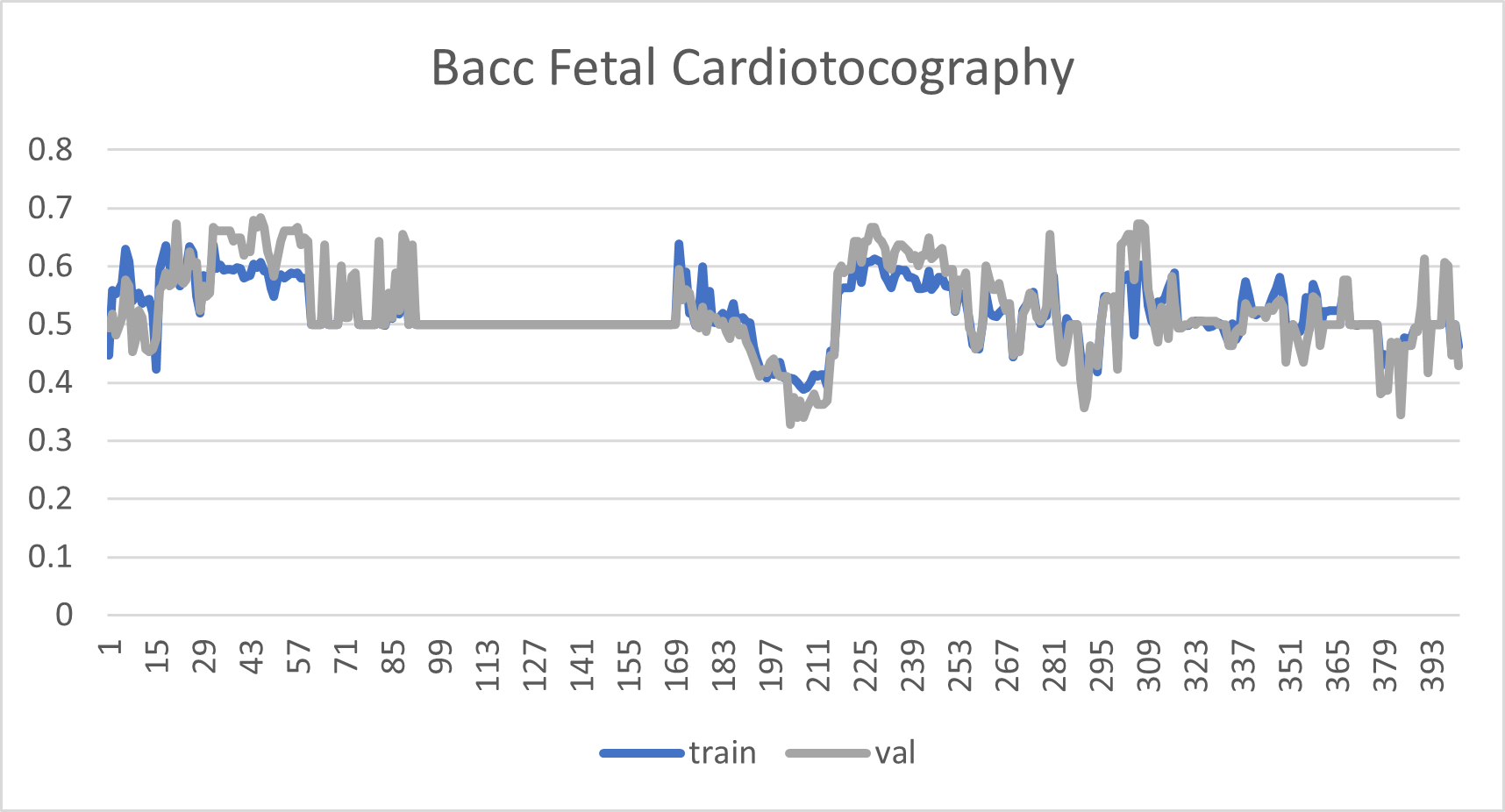}
        \caption{\added{Balanced Accuracy Curve on the \textbf{Fetal Cardiotocography} Dataset}}
        \label{fig:sub4}
    \end{subfigure}
    
    \caption{\added{Gradient descent BAcc curves for the four evaluated datasets: Synthetic, Hecktor, DLBCL, and Fetal.}}
    \label{fig:balanced_accuracy_curves}
\end{figure}

\section{Discussion}
According to our results, GA seems \replaced{indeed}{much} more suitable to train the DEBI-NN architecture than our proposed implementation of GD.\added{ It should be noted, however, that a broader hyperparameter range was explored for GD, which mechanically increases the likelihood of including less relevant configurations, leading to a higher standard deviation and a more pronounced difference with GA. Still, the most relevant comparison remains between the best-performing configurations of GA and GD (each optimized for its own hyperparameters) across the four datasets.}

This gap in performance can be explained by the highly non-convex optimization space created by DEBI-NN’s distance-encoded weight scheme. Indeed, in DEBI-NN, moving a single neuron in the 3D space simultaneously affects all its incident weights which is considered an inherent advantage of DEBI-NNs. However, this generates entangled gradient signals that can trap GD in suboptimal minima, \added{which could explain the high variability observed in the GD BAcc curve (Figure \ref{fig:balanced_accuracy_curves})}. In contrast, GA’s population-based mutation-and-crossover approach naturally allows exploring the spatial coordinate space and shows higher robustness to conflicting update directions.

Our findings should be read alongside recent work that trains spatially embedded networks with gradient descent and reports competitive results. Erb et al.~\cite{erb2025trainingneuralnetworksoptimizing} place each neuron as a single point in three-dimensional space with the layer coordinate fixed, define weights as the inverse of the Euclidean distance, learn a per-neuron excitatory or inhibitory sign, and train multilayer perceptrons and spiking networks with Adam on MNIST. Tohouri~\cite{Tohouri2025} uses single-point neurons with three free coordinates, maps distance to a bounded weight through a negative hyperbolic tangent that changes sign at a preferred distance, adds a regulariser pulling distances toward that value, and trains a bias-free network with identity activations for regression on the California Housing dataset. M\'esz\'aros et al.~\cite{meszaros2025spaceastime} use neuron positions to define synaptic delays rather than weights in recurrent spiking networks trained with event-based gradients on the Spiking Heidelberg Digits. These results do not contradict ours. None of these studies compares gradient descent against an evolutionary optimiser on the same architecture. Their reference point is a conventional non-spatial network, whereas ours is a genetic algorithm applied to an identical DEBI-NN. The architectures also differ in ways that change the optimisation landscape. DEBI-NN separates each neuron into a soma and an axon terminal, so a weight depends on two movable endpoints in different layers rather than on one point per neuron. The distance-to-weight mappings (Gaussian and linear in our case, inverse distance, bounded tangent or delay encoding in theirs) produce different gradient magnitudes and sign behaviour. The number of free coordinates per neuron ranges from two to six, and the stabilisation mechanisms (group normalisation and alternating endpoint updates here, distance regularisation or a learnable sign elsewhere) are not interchangeable. The datasets differ by two to three orders of magnitude in sample size, and the tasks span classification, regression and temporal classification. Erb et al.~\cite{erb2025trainingneuralnetworksoptimizing} also observe that their spatially embedded perceptrons stay below parameter-matched conventional perceptrons and attribute this to the same coupling between a neuron's position and all of its weights that we identify as the obstacle to backpropagation in DEBI-NN, and Tohouri~\cite{Tohouri2025} reports strong sensitivity to re-initialisation. The current literature therefore shows that spatially embedded networks can be trained by gradient descent under some combinations of encoding, mapping, regularisation and data scale, and that a genetic algorithm outperformed our gradient scheme under the DEBI-NN encoding on small medical datasets. Which of these design choices decides whether gradient-based training succeeds is still unknown, and a controlled study varying one factor at a time on a shared set of datasets would be needed to find out. Our conclusion on the limits of gradient descent applies to the DEBI-NN encoding, the datasets and the gradient scheme evaluated here, not to spatially embedded architectures in general.

Our work has several limitations. First, our GA implementation has benefited from several iterative refinements since its initial design and development, during which we were able to identify the most appropriate hyperparameter ranges to explore here. By comparison, our GD design and implementation is more recent and could still be suboptimal despite our efforts to explore a \replaced{larger}{reasonable} range of hyperparameters to optimize its performance for the comparison with GA. 

Second, our evaluation was restricted to \deleted{binary }classification tasks (in all three \added{real medical} datasets). \added{This is however the most frequent task in medical imaging and signal datasets. Nevertheless,} future work \replaced{will}{could} explore\deleted{ multi-class tasks and} regression tasks that are also of importance in medical \deleted{imaging} tasks (e.g., prediction of time-to-events endpoints such as survival). 

Thirdly, we did not explicitly investigate the impact of \added{varying} the number of samples \added{within each dataset}. However, we did evaluate three real-world scenarios with sample sizes ranging from n=85 to n=2126, observing consistent results across all of them.

Fourth, we implemented and evaluated the comparison between GA and GD in the context of a DEBI-NN defined in a 3D Euclidian space only. We hypothesize that reducing the dimensionality to 2D might not yield significant gains, as it severely restricts the spatial freedom required for neurons to find an optimal arrangement. Conversely, while a 4D DEBI-NN would offer larger degrees of freedom for spatial organization, it would mechanically increase the parameter count (8 per neuron instead of 6), thereby diminishing the architecture's advantage in low-data regimes. Furthermore, higher dimensions would significantly complicate visualization and interpretability of both the evolution of the architecture during the training, as well as the final trained architecture.. These advantages and disadvantages could be more or less important in the context of GA vs. GD, but this remains to be thoroughly investigated in a dedicated study on the number of dimensions for DEBI-NN implementation and is out of the scope of the present work.

Fifth, while high computational cost is often considered the primary drawback of genetic algorithms compared to gradient-based methods, we did not perform a formal comparison of training times in this study. The current DEBI-NN implementation is written in C++ and executed on CPU, and is not yet optimized for computational efficiency. Consequently, any comparison between GA and GD regarding training time or computational requirements would have little practical value at this stage. Developing a computationally efficient, and likely GPU-accelerated, version of DEBI-NN is an important task for future work.

Regarding future developments, we identify the implementation of a hybrid strategy as a promising research direction. Such an approach would combine the strengths of both learners: GA for global exploration as a first step, followed by GD for local fine-tuning as a second step. However, it is important to note that this specific scenario was not implemented in the current study and is left for future work. 

Finally, we explicitly state that the present investigations did not aim at providing a definitive answer regarding the potential value of DEBI-NN compared to other propositions for low-parameter count or spatially arranged neural network architectures. The sole objective of this study was to properly design a GD learner adapted for the DEBI-NN architecture in order to evaluate if the initial choice for the use of an evolutionary approach was indeed the best one.

\section{Conclusions}

In this study, we compared genetic algorithm (GA) and gradient descent (GD) for training the recently proposed DEBI-NN architecture, a neural network model based on the Euclidian spatial encoding of neurons in order to drastically reduce the number of trainable parameters. Across all evaluated datasets (synthetic and clinical), GA consistently outperformed GD in terms of balanced accuracy, sensitivity, and specificity, both approaches being optimized in terms of hyperparameters. Specifically, GA achieved 100\% balanced accuracy on the synthetic task (vs 83\% for GD) and consistently higher performance across the clinical datasets (DLBCL: 83\% vs 78\%; HECKTOR: 80\% vs 67\%; Fetal: 81\% vs 66\%). These findings confirm the limitations of gradient-based optimization for architectures with highly interdependent parameters, as is the case for DEBI-NN, where modifying a single neuron’s position affects all its associated weights. GA thus appears better suited to navigating the resulting non-convex landscape by leveraging its population-based exploratory dynamics.\deleted{ Nonetheless, these insights suggest that a hybrid approach combining the global search capabilities of GA with the local refinement potential of GD may offer promising directions for further investigations. }

\section{Data availability}

The executable enabling the reproduction of all experiments has been made publicly available.

In addition, the four datasets used for training, the full set of experimental parameters, and the corresponding results obtained for each configuration are provided \cite{Boukhari2025DEBINNData}.

These resources allow the experiments reported in this paper to be reproduced.

\section{Acknowledgments}
The present research was partly funded by the City of Brest (Brest Métropole Océane) and by La Ligue Contre le Cancer.

\bibliographystyle{elsarticle-num} 
\bibliography{GDvsGA.bib}

@inproceedings{erb2025trainingneuralnetworksoptimizing,
  author    = {Erb, Laura and Boccato, Tommaso and Vasilache, Alexandru and Becker, Juergen and Toschi, Nicola},
  title     = {Training Neural Networks by Optimizing Neuron Positions},
  booktitle = {Biomimetic and Biohybrid Systems. Living Machines 2025},
  series    = {Lecture Notes in Computer Science},
  volume    = {15582},
  publisher = {Springer},
  address   = {Cham},
  year      = {2026},
  doi       = {10.1007/978-3-032-07448-5_23},
  note      = {Preprint: arXiv:2506.13410}
}

@misc{Tohouri2025,
  author       = {Tohouri, Pascal},
  title        = {Training {3D} Spatially Embedded Neural Networks for Regression via Gradient Descent},
  year         = {2025},
  howpublished = {figshare preprint},
  doi          = {10.6084/m9.figshare.29588804},
  url          = {https://figshare.com/articles/preprint/Training_3D_Spatially_Embedded_Neural_Networks_for_Regression_via_Gradient_Descent/29588804},
  note         = {Version dated 2026-07-24, accessed 2026-09-16. Not peer reviewed.}
}

@misc{meszaros2025spaceastime,
  author        = {M{\'e}sz{\'a}ros, Bal{\'a}zs and Knight, James C. and Akarca, Danyal and Nowotny, Thomas},
  title         = {Space as Time Through Neuron Position Learning},
  year          = {2025},
  eprint        = {2511.01632},
  archivePrefix = {arXiv},
  primaryClass  = {cs.NE},
  note          = {v2, 26 January 2026}
}

@article{papp2023debi,
  title={DEBI-NN: Distance-encoding biomorphic-informational neural networks for minimizing the number of trainable parameters},
  author={Papp, Laszlo and Haberl, David and Ecsedi, Boglarka and Spielvogel, Clemens P and Krajnc, Denis and Grahovac, Marko and
others},
  journal={Neural Networks},
  volume={167},
  pages={517--532},
  year={2023},
  publisher={Elsevier}
}

@InProceedings{andrearczyk2022overview,
author="Andrearczyk, Vincent
and Oreiller, Valentin
and Abobakr, Moamen
and Akhavanallaf, Azadeh
and Balermpas, Panagiotis
and Boughdad, Sarah and
et al",
editor="Andrearczyk, Vincent
and Oreiller, Valentin
and Hatt, Mathieu
and Depeursinge, Adrien",
title="Overview of the HECKTOR Challenge at MICCAI 2022: Automatic Head and Neck Tumor Segmentation and Outcome Prediction in PET/CT",
booktitle="Head and Neck Tumor Segmentation and Outcome Prediction",
year="2023",
publisher="Springer Nature Switzerland",
address="Cham",
pages="1--30",
isbn="978-3-031-27420-6"
}

@article{ritter2022two,
  title={Two-year event-free survival prediction in DLBCL patients based on in vivo radiomics and clinical parameters},
  author={Ritter, Zsombor and Papp, L{\'a}szl{\'o} and Z{\'a}mb{\'o}, Katalin and T{\'o}th, Zolt{\'a}n and Dezs{\H{o}}, D{\'a}niel and Veres, D{\'a}niel S{\'a}ndor and others},
  journal={Frontiers in Oncology},
  volume={12},
  pages={820136},
  year={2022},
  publisher={Frontiers Media SA}
}

@article{zwanenburg2020imageIBSI,
  title={The image biomarker standardization initiative: standardized quantitative radiomics for high-throughput image-based phenotyping},
  author={Zwanenburg, Alex and Valli{\`e}res, Martin and Abdalah, Mahmoud A and Aerts, Hugo JWL and Andrearczyk, Vincent and Apte, Aditya and others},
  journal={Radiology},
  volume={295},
  number={2},
  pages={328--338},
  year={2020},
  publisher={Radiological Society of North America}
}

@article{llmscaling,
  title={Scaling laws for neural language models},
  author={Kaplan, Jared and McCandlish, Sam and Henighan, Tom and Brown, Tom B and Chess, Benjamin and Child, Rewon and others},
  journal={arXiv preprint arXiv:2001.08361},
  year={2020}
}

@inproceedings{sevilla2021parameter,
  title={Parameter counts in machine learning},
  author={Sevilla, Jaime and Villalobos, Pablo and Cer{\'o}n, Juan},
  booktitle={AI Alignment Forum},
  year={2021}
}

@article{celard2023survey,
  title={A survey on deep learning applied to medical images: from simple artificial neural networks to generative models},
  author={Celard, Pedro and Iglesias, Eva Lorenzo and Sorribes-Fdez, Jos{\'e} Manuel and Romero, Rub{\'e}n and Vieira, A Seara and Borrajo, Lourdes},
  journal={Neural Computing and Applications},
  volume={35},
  number={3},
  pages={2291--2323},
  year={2023},
  publisher={Springer}
}

@article{takahashi2024comparison,
  title={Comparison of vision transformers and convolutional neural networks in medical image analysis: a systematic review},
  author={Takahashi, Satoshi and Sakaguchi, Yusuke and Kouno, Nobuji and Takasawa, Ken and Ishizu, Kenichi and Akagi, Yu and others},
  journal={Journal of Medical Systems},
  volume={48},
  number={1},
  pages={84},
  year={2024},
  publisher={Springer}
}

@article{raghu2019transfusion,
  title={Transfusion: Understanding transfer learning for medical imaging},
  author={Raghu, Maithra and Zhang, Chiyuan and Kleinberg, Jon and Bengio, Samy},
  journal={Advances in neural information processing systems},
  volume={32},
  year={2019}
}

@article{he2024foundation,
  title={Foundation model for advancing healthcare: challenges, opportunities and future directions},
  author={He, Yuting and Huang, Fuxiang and Jiang, Xinrui and Nie, Yuxiang and Wang, Minghao and Wang, Jiguang and others},
  journal={IEEE Reviews in Biomedical Engineering},
  year={2024},
  publisher={IEEE}
}

@incollection{nayem2023few,
  title={Few shot learning for medical imaging: A comparative analysis of methodologies and formal mathematical framework},
  author={Nayem, Jannatul and Hasan, Sayed Sahriar and Amina, Noshin and Das, Bristy and Ali, Md Shahin and Ahsan, Md Manjurul and others},
  booktitle={Data Driven Approaches on Medical Imaging},
  pages={69--90},
  year={2023},
  publisher={Springer}
}

@misc{cardiotocography_193,
  author       = {Campos, D. and Bernardes, J.},
  title        = {{Cardiotocography}},
  year         = {2000},
  howpublished = {UCI Machine Learning Repository},
  note         = {{DOI}: https://doi.org/10.24432/C51S4N}
}

@article{ECSEDI2025100008,
title = {Impact of regularization in optimizing distance-encoding biomorphic-informational neural networks for small nuclear medicine datasets},
journal = {EANM Innovation},
volume = {1},
pages = {100008},
year = {2025},
issn = {3051-2913},
doi = {https://doi.org/10.1016/j.eanmi.2025.100008},
author = {Boglarka Ecsedi and Amine Boukhari and Clemens P. Spielvogel and David Haberl and Zsombor Ritter and Ralph A. Bundschuh and Constantin Lapa and Marcus Hacker and Mathieu Hatt and Laszlo Papp},
}

@book{papp_2025_17224628,
  author       = {Papp, László},
  title        = {Mastering Distance-Encoding Biomorphic Neural
                   Networks – The DEBI-NN Handbook
                  },
  publisher    = {Zenodo},
  year         = 2025,
  month        = oct,
  doi          = {10.5281/zenodo.17224628},
  url          = {https://doi.org/10.5281/zenodo.17224628},
}

@misc{wu2018groupnormalization,
      title={Group Normalization}, 
      author={Yuxin Wu and Kaiming He},
      year={2018},
      eprint={1803.08494},
      archivePrefix={arXiv},
      primaryClass={cs.CV},
      url={https://arxiv.org/abs/1803.08494}, 
}

@misc{Boukhari2025DEBINNData,
  doi = {10.17632/8KWN35PPCD},
  url = {https://data.mendeley.com/datasets/8kwn35ppcd},
  author = {boukhari, amine and ecsedi, boglárka  and papp, laszlo and hatt, mathieu},
  title = {DEBI-NN: Genetic algorithm vs. gradient descent data},
  publisher = {Mendeley Data},
  year = {2025}
}

@article{Hecktor21,
title = {Automatic Head and Neck Tumor segmentation and outcome prediction relying on FDG-PET/CT images: Findings from the second edition of the HECKTOR challenge},
journal = {Medical Image Analysis},
volume = {90},
pages = {102972},
year = {2023},
issn = {1361-8415},
doi = {https://doi.org/10.1016/j.media.2023.102972},
url = {https://www.sciencedirect.com/science/article/pii/S1361841523002323},
author = {Vincent Andrearczyk and Valentin Oreiller and Sarah Boughdad and Catherine Cheze {Le Rest} and Olena Tankyevych and Hesham Elhalawani and Mario Jreige and John O. Prior and Martin Vallières and Dimitris Visvikis and Mathieu Hatt and Adrien Depeursinge}
}



\end{document}